%% file: main.tex
\documentclass[10pt,twocolumn,letterpaper]{article}

\usepackage{cvpr}
\usepackage{times}
\usepackage{epsfig}
\usepackage{graphicx}
\usepackage{amsmath}
\usepackage{amssymb}
\input{math_commands.tex}

\usepackage[utf8]{inputenc} % allow utf-8 input
\usepackage[T1]{fontenc}    % use 8-bit T1 fonts
\usepackage{url}            % simple URL typesetting
\usepackage{booktabs}       % professional-quality tables
\usepackage{amsfonts}       % blackboard math symbols
\usepackage{nicefrac}       % compact symbols for 1/2, etc.
\usepackage{microtype}      % microtypography
\usepackage{graphicx}
\usepackage{enumerate}
\usepackage{algpseudocode,algorithm,algorithmicx}  
\usepackage{caption}% http://ctan.org/pkg/caption

\usepackage[pagebackref=true,breaklinks=true,letterpaper=true,colorlinks,bookmarks=false]{hyperref}

\usepackage{color,soul}
\usepackage[nomargin,inline,marginclue,draft]{fixme}

\usepackage{amsmath,amssymb,bm}

\title{Representation Flow for Action Recognition}

\author{AJ Piergiovanni and Michael S. Ryoo\\
Department of Computer Science, Indiana University, Bloomington, IN 47408 \\
\texttt{\{ajpiergi,mryoo\}@indiana.edu}\\
}

\cvprfinalcopy % *** Uncomment this line for the final submission

\def\cvprPaperID{4973} % *** Enter the CVPR Paper ID here

\ifcvprfinal\fi
\begin{document}

\maketitle

\begin{abstract}

In this paper, we propose a convolutional layer inspired by optical flow algorithms to learn motion representations. Our \emph{representation flow} layer is a fully-differentiable layer designed to capture the `flow' of any representation channel within a convolutional neural network for action recognition. Its parameters for iterative flow optimization are learned in an end-to-end fashion together with the other CNN model parameters, maximizing the action recognition performance. Furthermore, we newly introduce the concept of learning `flow of flow' representations by stacking multiple representation flow layers.
We conducted extensive experimental evaluations, confirming its advantages over previous recognition models using traditional optical flows in both computational speed and performance. The code is publicly available. \footnote{Code/models available here: \href{https://piergiaj.github.io/rep-flow-site/}{https://piergiaj.github.io/rep-flow-site/}}
\end{abstract}

\section{Introduction}

Activity recognition is an important problem in computer vision with many societal applications including surveillance, robot perception, smart environment/city, and more. Use of video convolutional neural networks (CNNs) have become the standard method for this task, as they can learn more optimal representations for the problem. Two-stream networks \cite{simonyan2014two}, taking both RGB frames and optical flow as input, provide state-of-the-art results and have been extremely popular. 3-D spatio-temporal CNN models, e.g., I3D \cite{carreira2017quo}, with XYT convolutions also found that such two-stream design (RGB + optical flow) increases their accuracy. Abstracting both appearance information and explicit motion flow benefits the recognition.

%Even large, 3D CNNs \cite{carreira2017quo}, require optical flow to achieve state-of-the-art performance.

However, optical flow is expensive to compute. It often requires hundreds of optimization iterations every frame, and causes learning of two separate CNN streams (i.e., RGB-stream and flow-stream). This requires significant computation cost and a great increase in the number of model parameters to learn. Further, this means that the model needs to compute optical flow every frame even during inference and run two parallel CNNs, limiting its real-time applications.

\begin{figure}
    \centering
    \includegraphics[width=0.48\textwidth]{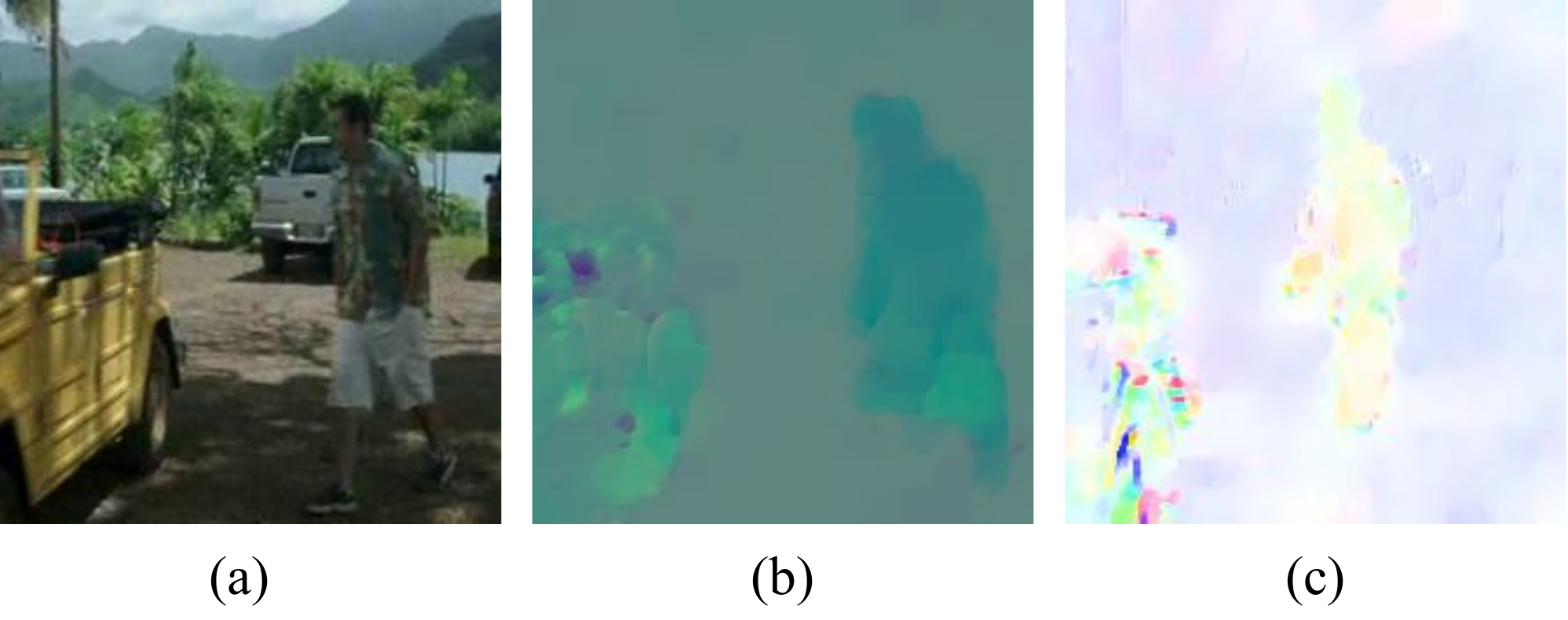}
    \caption{Comparing the results of (b) TVL-1 and (c) our learned flow when applied to RGB images. Our layer is able to capture similar motion information to TVL-1. However, compared to TVL-1, our layer is faster, is learnable, and can be applied directly to any intermediate CNN feature maps. With the representation flow layer, optical flow pre-extraction is no longer needed and a single-stream CNN design becomes possible.}
    \label{fig:flow-comapre}
\end{figure}

There were previous works to learn representations capturing motion information without using optical flow as input, such as motion feature networks \cite{lee2018motion} and ActionFlowNet \cite{ng2018actionflownet}. However, although they were more advantageous in terms of the number of model parameters and computation speed, they suffered from inferior performance compared to two-stream models on public datasets such as Kinetics \cite{kay2017kinetics} and HMDB \cite{hmdb}. We hypothesize that the iterative optimization performed by optical flow methods produces an important feature that other methods fail to capture.

In this paper, we propose a CNN layer inspired by optical flow algorithms to learn motion representations for action recognition without having to compute optical flow.
Our \emph{representation flow} layer is a fully-differentiable layer designed to capture `flow' of any representation channels within the model.
%(i.e., not just RGB).
Its parameters for iterative flow optimization are learned together with other model parameters, maximizing the action recognition performance.
This is also done without having/training multiple network streams, reducing the number of parameters in the model.
Further, we newly introduce the concept of learning `flow of flow' representations by stacking multiple representation flow layers.
We conduct extensive action classification experimental evaluation of where to compute optical flow and various hyperparameters, learning parameters, and fusion techniques.

Our contribution is the introduction of a new differentiable CNN layer that unrolls the iterations of the TV-L1 optical flow method. This allows for learning of the optical flow parameters, application to any CNN feature maps (i.e., intermediate representations), and lower computational cost while maintaining performance.

\section{Related Works}
Capturing motion and temporal information has been studied for activity recognition. Early, hand-crafted approaches such as dense trajectories \cite{wang2011action} captured motion information by tracking points through time. Many algorithms have been developed to compute optical flow as a way to capture motion in video \cite{fortun2015optical}. Other works have explored learning the ordering of frames to summarize a video in a single `dynamic image' used for activity recognition \cite{bilen2016dynamic}.

Convolutional neural networks (CNNs) have been applied to activity recognition. Initial approaches explored methods to combine temporal information based on pooling or temporal convolution \cite{karpathy2014large,ng2015beyond}. Other works have explored using attention to capture sub-events of activities \cite{piergiovanni2017learning}. Two-stream networks have been very popular: they take input of a single RGB frame (captures appearance information) and a stack of optical flow frames (captures motion information). Often, the two network streams of the model are separately trained and the final predictions are averaged together \cite{simonyan2014two}. There were other two-stream CNN works exploring different ways to `fuse' or combine the motion CNN with the appearance CNN \cite{feichtenhofer2016convolutional,feichtenhofer2016spatiotemporal}.
There were also large 3D XYT CNNs learning spatio-temporal patterns \cite{xie2017rethinking,carreira2017quo}, enabled by large video datasets such as Kinetics \cite{kay2017kinetics}.
%Large video datasets, such as Kinetics \cite{kay2017kinetics}, enabled the training of large 3D CNNs which are able to learn spatio-temporal patterns \cite{carreira2017quo,xie2017rethinking}.
However, these approaches still rely on optical flow input to maximize their accuracies.

While optical flow is known to be an important feature, flows optimized for activity recognition are often different from the true optical flow \cite{sevilla2018integration}, suggesting that end-to-end learning of motion representations is beneficial. Recently, there have been works on learning such motion representations using convolutional models. Fan et al. \cite{fan2018end} implemented the TV-L1 method using deep learning libraries to increase its computational speed and allow for learning some parameters. The result was fed to a two-stream CNN for the recognition. Several works explored learning a CNN to predict optical flow, which also can be used for action recognition \cite{dosovitskiy2015flownet,gao2017im2flow,hui2018liteflownet,ng2018actionflownet,sun2018optical}. Lee et al. \cite{lee2018motion} shifted features from sequential frames to capture motion in a non-iterative fashion. Sun et al. \cite{sun2018optical} proposed an optical flow guided feature (OFF) by computing the gradients of representations and temporal differences, but it lacked the iterative optimization necessary for accurate flow computation. Further, it requires a three-stream model taking RGB, optical flow, and RGB differences to achieve state-of-the-art performance.

%Our proposed representation flow layer relies only on the RGB input, is significantly faster than optical flow methods and outperforms existing motion representation methods.

Unlike prior works, our proposed model with \emph{representation flow} layers relies only on RGB input, learning far fewer parameters while correctly representing motion with the iterative optimization. It is significantly faster than the video CNNs requiring optical flow input, while still performing as good as or even better than the two-stream models. It clearly outperforms existing motion representation methods including TVNet \cite{fan2018end} and OFF \cite{sun2018optical} in both speed and accuracy, which we experimentally confirm.

\section{Approach}
Our method is a fully-differentiable convolutional layer inspired by optical flow algorithms. Unlike traditional optical flow methods, all the parameters of our method can be learned end-to-end, maximizing action recognition performance. Furthermore, our layer is designed to compute the `flow' of any representation channels, instead of limiting its input to be traditional RGB frames.

\subsection{Review of Optical Flow Methods}
Before describing our layer, we briefly review how optical flow is computed. Optical flow methods are based on the brightness consistency assumption. That is, given sequential images $I_1,I_2$, a point $x,y$ in $I_1$ is located at $x+\Delta x, y+\Delta y$ in $I_2$, or $I_1(x,y) = I_2(x+\Delta x, y+\Delta y)$. These methods assume small movements between frames, so this can be approximated with a Taylor series: $I_2 = I_1 + \frac{\delta I}{\delta x}\Delta x + \frac{\delta I}{\delta y}\Delta y$, where $\bm{u}=[\Delta x, \Delta y]$. These equations are solved for $\bm{u}$ to obtain the flow, but can only be approximated due to the two unknowns.

The standard, variational methods for approximating optical flow (e.g., Brox \cite{brox2004high} and TV-L1 \cite{zach2007duality} methods) take sequential images $I_1,I_2$ as input. Variational optical flow methods estimate the flow field, $\bm{u}$, using an iterative optimization method. The tensor $\bm{u}\in\mathcal{R}^{2\times W\times H}$ is the $x$ and $y$ directional flow for every location in the image. Taking two sequential images as input, $I_1,I_2$, the methods first compute the gradient in both $x$ and $y$ directions: $\nabla I_2$. The initial flow is set to 0, $\bm{u}=0$. Then $\rho$, which captures the motion residual between two frames, based on the current flow estimate $\bm{u}$, can be computed. For efficiency, the constant part of $\rho$, $\rho_c$ is pre-computed:
\begin{equation}
    \rho_c = I_2 - \nabla_x I_2 \cdot \bm{u}_x  - \nabla_y I_2 \cdot \bm{u}_y - I_1
\end{equation}

The iterative optimization is then performed, each updating $\bm{u}$:
\begin{alignat}{4}
    \rho &= \rho_c + \nabla_x I_2\cdot\bm{u}_x + \nabla_y I_2\cdot\bm{u}_y \label{eq:rho}\\
    \bm{v} &= \begin{cases}
        \bm{u} + \lambda\theta \nabla I_2 & \rho < -\lambda\theta |\nabla I_2|^2\\
        \bm{u} - \lambda\theta \nabla I_2 & \rho > \lambda\theta |\nabla I_2|^2\\
        \bm{u} - \rho \frac{\nabla I_2}{|I_2|^2}  & \text{otherwise}
    \end{cases}\\
    \bm{u} &= \bm{v} + \theta\cdot \text{divergence}(\bm{p})\\
    \bm{p} &= \frac{\bm{p}+\frac{\tau}{\theta}\nabla\bm{u}}{1+\frac{\tau}{\theta}|\nabla\bm{u}|}\label{eq:p}
\end{alignat}

Here $\theta$ controls the weight of the TV-L1 regularization term, $\lambda$ controls the smoothness of the output and $\tau$ controls the time-step. These hyperparameters are manually set. $\bm{p}$ is the dual vector fields, which are used to minimize the energy. The divergence of $\bm{p}$, or backward difference, is computed as:
\begin{equation}
\label{eq:divergence}
 \text{divergence}(\bm{p})=\bm{p}_{x,i,j}-\bm{p}_{x,i-1,j} + \bm{p}_{y,i,j}-\bm{p}_{y,i,j-1}
\end{equation}
where $\bm{p}_x$ is the $x$ direction and $\bm{p}_y$ is the $y$ direction, and $\bm{p}$ contains all the spatial locations in the image.

The goal is to minimize the total variational energy:
\begin{equation}
    E = |\nabla \bm{u}| + \lambda|\nabla I_1 * u + I_1 - I_2|
\end{equation}

Approaches run this iterative optimization for multiple input scales, from small to large, and use the previous flow estimate $\bm{u}$ to warp $I_2$ at the larger scale, providing a coarse-to-fine optical flow estimation. These standard approaches require multiple scales and warpings to obtain a good flow estimate, taking thousands of iterations.

\subsection{Representation Flow Layer}
Inspired by the optical flow algorithm, we design a fully-differentiable, learnable, convolutional representation flow layer by extending the general algorithm outlined above. The main differences are that (i) we allow the layer to capture flow of any CNN feature map, and that (ii) we learn its parameters including $\theta$, $\lambda$, and $\tau$ as well as the divergence weights. We also make several key changes to reduce computation time: (1) we only use a single scale, (2) we do not perform any warping, and (3) we compute the flow on a CNN tensor with a smaller spatial size. Multiple scale and warping are computationally expensive, each requiring many iterations. By learning the flow parameters, we can eliminate the need for these additional steps. Our method is applied on lower resolution CNN feature maps, instead of the RGB input, and is trained in an end-to-end fashion. This not only benefits its speed, but also allows the model to learn a motion representation optimized for activity recognition.

We note that the brightness consistency assumption can similarly be applied to CNN feature maps. Instead of capturing pixel brightness, we capture feature value consistency. This same assumption holds as CNNs are designed to be spatially invariant; i.e., they  produce roughly the same feature value for the same object as it moves.

Given the input $F_1,F_2$, a single channel from sequential CNN feature maps (or input image), we compute the feature-map-gradient by convolving the input feature maps with the Sobel filter:
\begin{equation}
\label{eq:img-grad}
    \nabla F_{2x} = \begin{bmatrix}
    1 & 0 & -1\\
    2 & 0 & -2\\
    1 & 0 & -1
\end{bmatrix}
 * F_2, ~~  \nabla F_{2y} = \begin{bmatrix}
    1 & 2 & 1\\
    0 & 0 & 0\\
    -1 & -2 & -1
\end{bmatrix}
 * F_2
\end{equation}

\begin{figure}
    \centering
      \includegraphics[width=\linewidth]{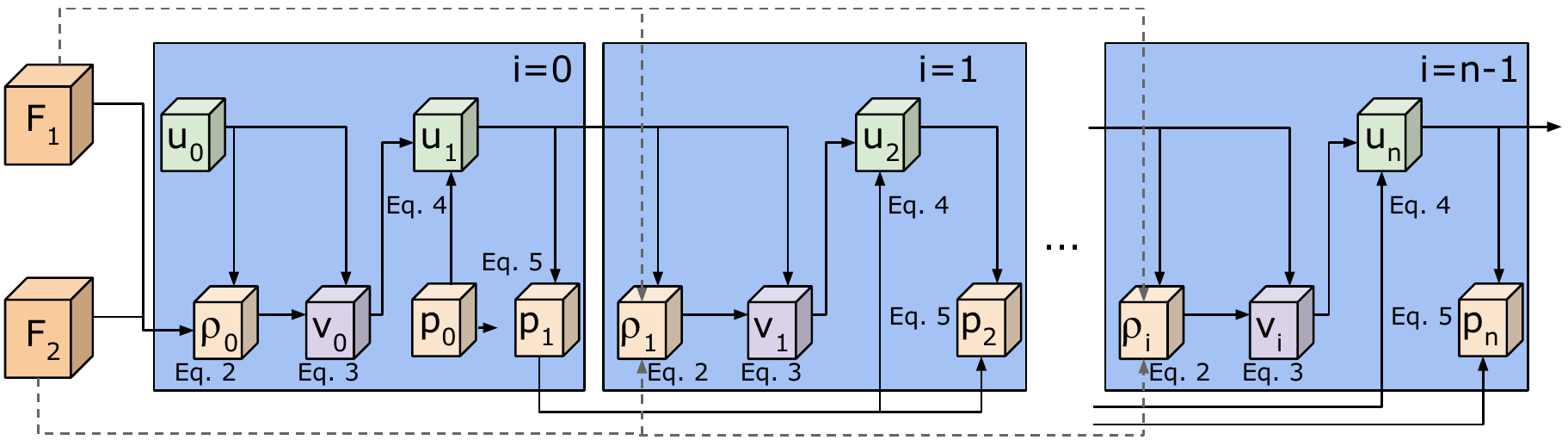}
      \caption{Illustration of our flow layer. It unrolls the iterations of the TV-L1 algorithm as a sequence of tensor operations, while sharing parameters across the iterations.}
      \label{fig:flow-layer}
\end{figure}

We set $\bm{u}=0, \bm{p}=0$ initially, each having width and height matching the input, then we can compute $\rho_c = F_2 - F_1$. Next, following Algorithm~\ref{alg:flow}, we repeatedly apply the operations in Eqs.~\ref{eq:rho}-\ref{eq:p} for a fixed number of iterations to enable the iterative optimization. 
%Next, we run the iterative optimization for a fixed number of iterations, following Eqs.~\ref{eq:rho}-\ref{eq:p}. The main idea is that we are able to make the representations and the parameters in Eqs.~\ref{eq:rho}-\ref{eq:p} differentiable so that the gradients can backpropagate through them even with their iterative applications.
To compute the divergence, we zero-pad $\bm{p}$ on the first column (x-direction) or row (y-direction) then convolve it with weights, $w_x,w_y$ to compute Eq.~\ref{eq:divergence}:
\begin{equation}
   \text{divergence}(\bm{p}) = \bm{p}_x * w_x + \bm{p}_y * w_y
\end{equation}
where initially $w_x=\begin{bmatrix}-1&1\end{bmatrix}$ and $w_y=\begin{bmatrix}-1\\1\end{bmatrix}$. Note that these parameters are also differentiable and can be learned with backpropagation. We compute $\nabla\bm{u}$ as
\begin{equation}
    \nabla\bm{u}_x = \begin{bmatrix}
    1 & 0 & -1\\
    2 & 0 & -2\\
    1 & 0 & -1
\end{bmatrix}
 * \bm{u}_x, ~~  \nabla \bm{u}_{y} = \begin{bmatrix}
    1 & 2 & 1\\
    0 & 0 & 0\\
    -1 & -2 & -1
\end{bmatrix}
 * \bm{u}_y
\end{equation}

\vspace{-3pt}
\paragraph{Representation Flow within a CNN}
\begin{figure*}
    \centering
      \includegraphics[width=0.95\textwidth]{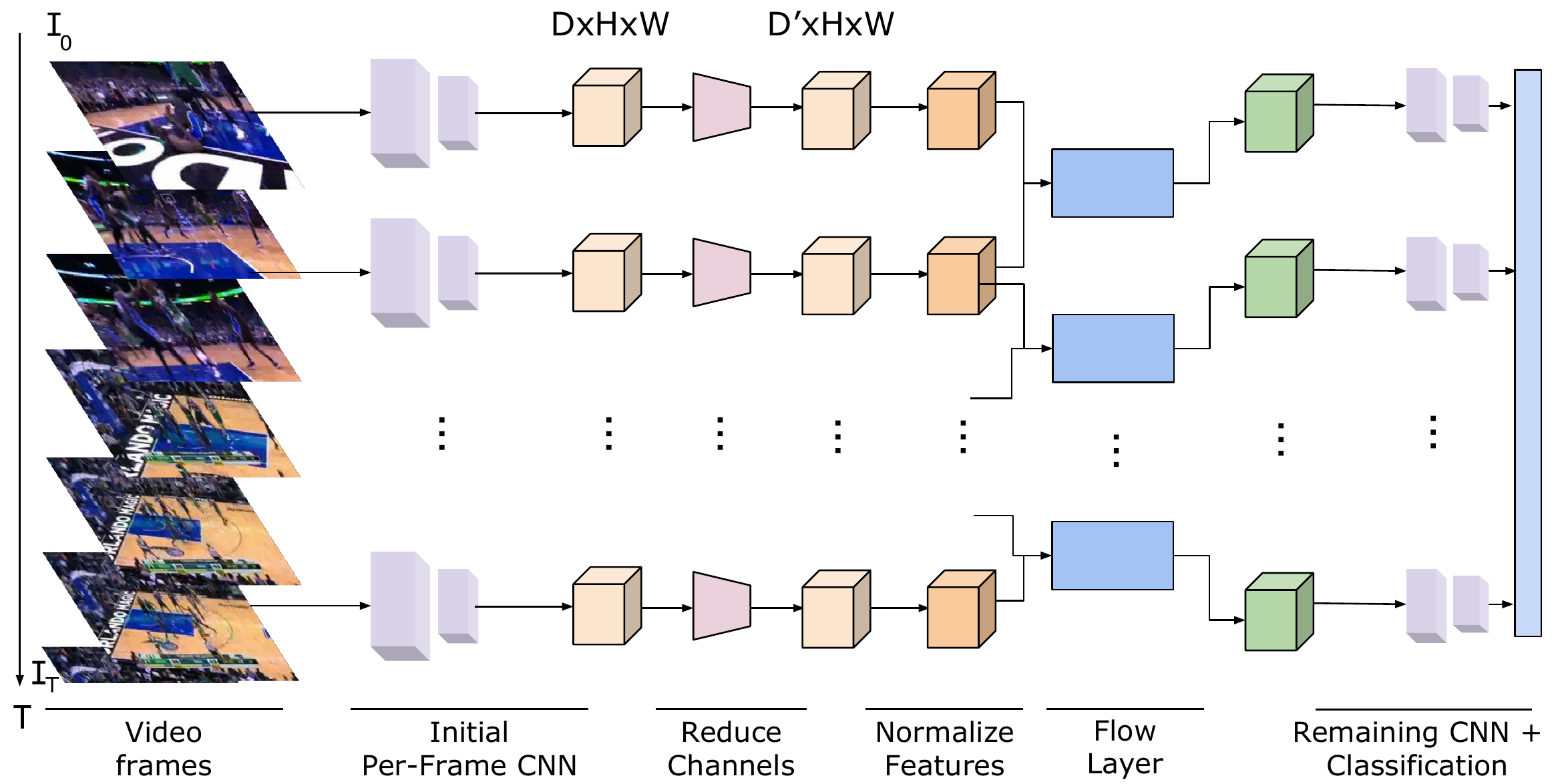}
      \caption{Illustration of a video-CNN with our representation flow layer. The CNN computes intermediate feature maps, and sequential feature maps are used as input to the flow layer. The outputs of the flow layer are used for prediction.}
      \label{fig:flow-method}
\end{figure*}

Algorithm~\ref{alg:flow} and Fig.~\ref{fig:flow-layer} describe the process of our representation flow layer.
%We can compute the representation flow for a given input following Algorithm~\ref{alg:flow}.
Our flow layer with multiple iterations could also be interpreted as having a sequence of convolutional layers sharing parameters (i.e., each blue box in Fig.~\ref{fig:flow-layer}), with each layer's behavior dependent on its previous layer. As a result of this formulation, the layer becomes fully differentiable and allows for the learning of all parameters, including ($\tau,\lambda,\theta$) and the divergence weights $(w_x, w_y)$. This enables our learned representation flow layer to be optimized for its task (i.e., action recognition).

\begin{algorithm}  
  \caption{Method for the representation flow layer
    \label{alg:flow}}  
  \begin{algorithmic}  
    \Function{RepresentationFlow}{$F_1$, $F_2$}
    \State $\bm{u} = 0, \bm{p} = 0$
    \State Compute image/feature map gradients (Eq. \ref{eq:img-grad})
    \State $\rho_c = F_2 - F_1$
    \For{$n$ iterations}
       \State $\rho = \rho_c + \nabla_x F_2\cdot\bm{u}_x + \nabla_y F_2\cdot\bm{u}_y$
        \State $\bm{v} = \begin{cases}
        \bm{u} + \lambda\theta \nabla F_2 & \rho < -\lambda\theta |\nabla F_2|^2\\
        \bm{u} - \lambda\theta \nabla F_2 & \rho > \lambda\theta |\nabla F_2|^2\\
        \bm{u} - \rho \frac{\nabla F_2}{|F_2|^2}  & \text{otherwise}
    \end{cases}$
        \State $\bm{u} = \bm{v} + \theta\cdot \text{divergence}(\bm{p})$
        \State $\bm{p} = \frac{\bm{p}+\frac{\tau}{\theta}\nabla\bm{u}}{1+\frac{\tau}{\theta}|\nabla\bm{u}|}$
    \EndFor
    \State \Return $\bm{u}$
    \EndFunction
  \end{algorithmic}  
\end{algorithm}

\vspace{-3pt}
\paragraph{Computing Flow-of-Flow}
Standard optical flow algorithms compute the flow for two sequential images. An optical flow image contains information about the direction and magnitude of the motion. Applying the flow algorithm directly on two flow images means that we are tracking pixels/locations showing similar motion in two consecutive frames.
%As such, applying the optical flow algorithm directly on two flow images is often not very informative.
In practice, this typically leads to a worse performance due to inconsistent optical flow results and non-rigid motion.
On the other hand, our representation flow layer is `learned' from the data, and is able to suppress such inconsistency and better abstract/represent motion by having multiple regular convolutional layers between the flow layers. Fig. \ref{fig:fof-model} illustrates such design, which we confirm its benefits in the experiment section. By stacking multiple representation flow layers, our model is able to capture longer temporal intervals and consider locations with motion consistency.

%Thus applying the optical flow algorithm directly on two flow images does not make sense. However, since we are computing representation flow, we can apply our flow layer, a convolutional layer, then a flow layer (Fig. \ref{fig:fof-model}), allowing our model to capture longer-term motion information.

CNN feature maps may have hundreds or thousands of channels and our representation flow layer computes the flow for each channel, which can take significant time and memory. To address this, we apply a convolutional layer to reduce the number of channels from $C$ to $C'$ before the flow layer (note that $C'$ is still significantly more than traditional optical flow algorithms, which were only applied to single-channel, greyscale images). For numerical stability, we normalize this feature map to be in $[0,255]$, matching standard image values. We found that the CNN features were quite small on average ($<0.5$) and the TVL-1 algorithm default hyperparameters are designed for standard images values in $[0,255]$, thus we found this normalization step important. Using the normalized feature, we compute the flow and stack the $x$ and $y$ flows, resulting in $2C'$ channels. Finally, we apply another convolutional layer to convert from $2C'$ channels to $C$ channels. This is passed to the remaining CNN layers for the prediction.
%Then the remaining CNN layers are applied to predict the class.
We average predictions from many frames to classify each video, as shown in Fig. \ref{fig:flow-method}.

\subsection{Activity Recognition Model}
We place the representation flow layer inside a standard activity recognition model taking a $T\times C\times W\times H$ tensor as input to a CNN. Here, $C$ is 3 as our model uses direct RGB frames as an input. $T$ is the number of frames the model processes, and $W$ and $H$ are the spatial dimensions. The CNN outputs a prediction per-timestep and these are temporally averaged to produce a probability for each class. The model is trained to minimize cross-entropy:
\begin{equation}
L(v,c) = -\sum_i^K (c==i) \log(p_i)
\end{equation}
where $p=M(v)$, $v$ is the video, the function $M$ is the classification CNN and $c$ represents which of the $K$ classes $v$ belongs. That is, the parameters in our flow layers are trained together with the other layers, so that it maximizes the final classification accuracy.

\section{Experiments}

\vspace{-3pt}
\paragraph{Implementation details} We implemented our representation flow layer in PyTorch and our code and models are available. As training CNNs on videos is computationally expensive, we used a subset of the Kinetics dataset \cite{kay2017kinetics} with 100k videos from 150 classes: Tiny-Kinetics. This allowed testing many models more quickly, while still having sufficient data to train large CNNs. For most experiments, we used ResNet-34 \cite{he2016deep} with input of size $16\times 112\times 112$ (i.e., 16 frames with spatial size of 112). To further reduce the computation time for many studies, we used this smaller input, which reduces performance, but allowed us to use larger batch sizes and run many experiments more quickly. Our final models are trained on standard $224\times 224$ images. Check Appendix for specific training details.

% \paragraph{Baselines} We compare to many baselines, such as 2D or 3D RGB/Flow/Two-stream CNNs and previous approaches to learning motion representations (e.g. []). 

\begin{figure}
    \centering
    \includegraphics[width=\linewidth]{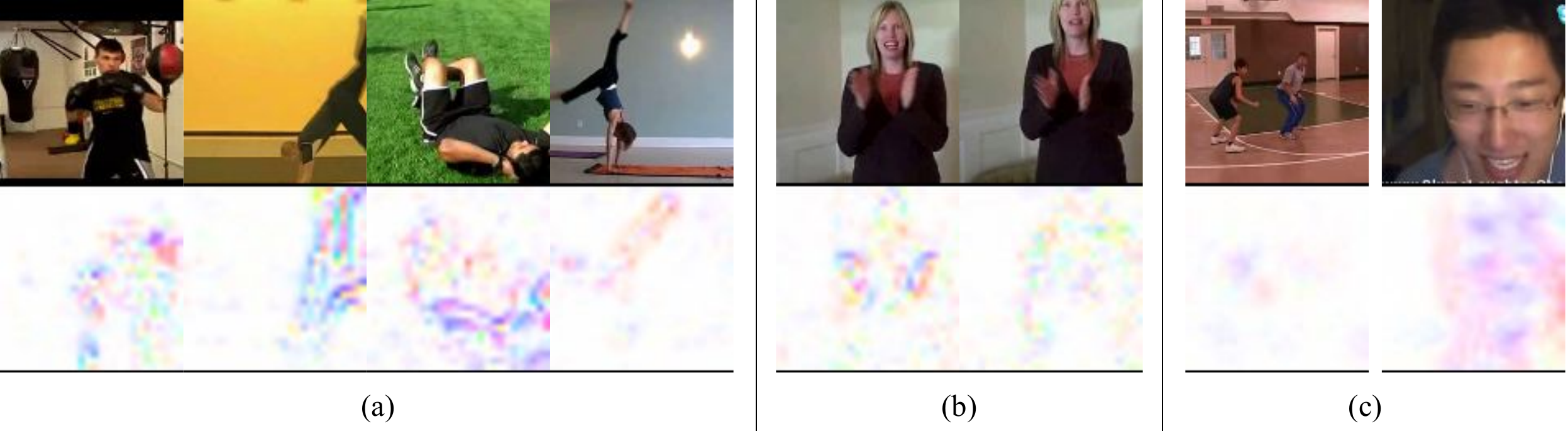}
    \caption{Visualization of learned representation flows. Note these are after the 3rd residual block and are low-resolution (28x28). (a) Examples of rep. flow for various activities. (b) Example of different channels capturing different motions: (left) hands (right) other motion. (c) Flow-of-flow is an acceleration-like feature with smoother, more abstract motion patterns. }
    \label{fig:vis-flow}
    \vspace{-10pt}
\end{figure}

\vspace{-3pt}
\paragraph{Where to compute flow?} To determine where in the network to compute the flow, we compare applying our flow layer on the RGB input, after the first conv. layer, and after the each of the 5 residual blocks. The results are shown in Table \ref{tab:where}. We find that computing the flow on the input provides poor performance, similar to the performance of the flow-only networks, but there is a significant jump after even 1 layer, suggesting that computing the flow of a feature is beneficial, capturing both the appearance and motion information. However, after 4 layers, the performance begins to decline as the spatial information is too abstracted/compressed (due to pooling and large spatial receptive field size), and sequential features become very similar, containing less motion information. Note that our HMDB performance in this table is quite low compared to state-of-the-art methods due to being trained from scratch using few frames and low spatial resolution ($112\times 112$). For the following experiments, unless otherwise noted, we apply the layer after the 3rd residual block. In Fig. \ref{fig:vis-flow}, we visualize the learned motion representations computer after block 3.

\begin{table}
  \caption{Computing the optical flow representation after various number of CNN layers. Results are video classification accuracy on our Tiny-Kinetics and LowRes-HMDB51 datasets using 100 iterations to compute the flow representation.}
  \small
  \label{tab:where}
  \centering
  \begin{tabular}{lcc}
    \toprule
         &  Tiny-Kinetics & LowRes-HMDB \\
    \midrule
    RGB CNN         & 55.2  & 35.5 \\
    Flow CNN        & 35.4  & 37.5 \\
    Two-Stream CNN  & 57.6  & 41.5 \\
    Flow Layer on RGB Input     & 37.4  & 40.5   \\
    After Block 1   & 52.4  & 42.6   \\
    After Block 2   & 57.4  & 44.5   \\
    After Block 3   & {\bf 59.4}  & {\bf 45.4}   \\
    After Block 4   & 52.1  & 43.5   \\
    After Block 5   & 50.3  & 42.2   \\
    \bottomrule
  \end{tabular}
\end{table}

\vspace{-3pt}
\paragraph{What to learn?} As our method is fully differentiable, we can learn any of the parameters, such as the kernels used to compute image gradients, the kernels for the divergence computation and even $\tau,\lambda,\theta$.  In Table \ref{tab:learn}, we compare the effects of learning different parameters. We find that learning the Sobel kernel values reduces performance due to noisy gradients particularly when the batch size is limited, but learning the divergence and $\tau,\lambda,\theta$ is beneficial.

\begin{table}
  \caption{Comparison of learning different parameters. The flow was computed after Block 3 using 100 iterations.}
  \label{tab:learn}
  \centering
  \begin{tabular}{lcc}
    \toprule
                                     &  Tiny-Kinetics & LowRes-HMDB \\
    \midrule
    None (all fixed)                   & 59.4  & 45.4   \\
    Sobel kernels                      & 58.5  & 43.5 \\
    Divergence ($w_x,w_y$)                 & 60.2  &  46.4 \\
    $\tau,\lambda,\theta$              & 59.9  & 46.2 \\
    All                                & 59.2  & 46.2  \\
    Divergence + $\tau,\lambda,\theta$   & {\bf 60.7}  & {\bf 46.8} \\
    \bottomrule
  \end{tabular}
\end{table}

\vspace{-3pt}
\paragraph{How many iterations for flow?} To confirm that the iterations are important and determine how many we need, we experiment with various numbers of iterations. We compare the number of iterations needed for both learning (divergence+$\tau,\lambda,\theta$) and not learning parameters. The flow is computed after 3 residual blocks. The results are shown in Table \ref{tab:iters}. We find that learning provides better performance with fewer iterations (similar to the finding in \cite{fan2018end}), and that iteratively computing the feature is important. We use 10 or 20 iterations in the remaining experiments as they provide good performance and are fast.

\begin{table}
  \caption{Effect of the number of iterations on our Tiny-Kinetics dataset for learning and not learning.}
  \label{tab:iters}
  \centering
  \begin{tabular}{lcc}
    \toprule
                   &  Not learned & Learned \\
    \midrule
    1 iteration       & 46.7  & 49.5   \\
    5 iterations      & 51.3  & 55.4  \\
    10 iterations     & 52.4  & 59.4  \\
    20 iterations     & 53.6  & {\bf 60.7}   \\
    50 iterations     & 59.2  & {\bf 60.9}   \\
    100 iterations    & 59.4  & 60.7 \\
    \bottomrule
  \end{tabular}
\end{table}

\begin{figure}
    \centering
      \includegraphics[width=\linewidth]{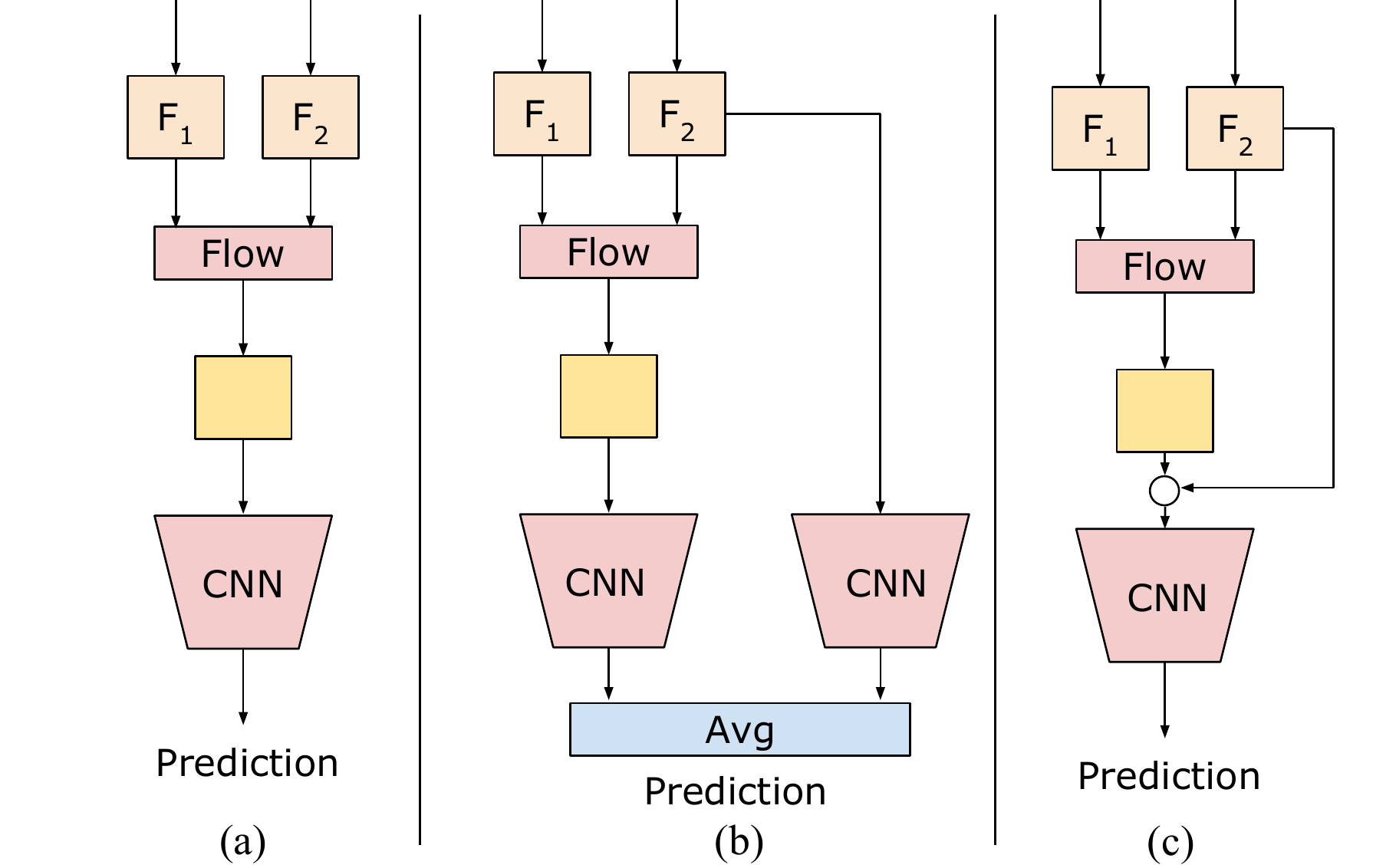}
      \caption{Different approaches to fusing RGB and flow information. {\bf (a)} No fusion {\bf (b)} Late fusion {\bf (c)} The circle represents elementwise addition/multiplication or concatenation. We experimentally find that no fusion performs comparably, when applied to after 3rd residual block.}
      \label{fig:fuse-method}
\end{figure}

\vspace{-3pt}
\paragraph{Two-stream fusion?} Two-stream CNNs fusing both RGB and optical flow features has been heavily studied \cite{simonyan2014two,feichtenhofer2016convolutional}. Based on these works, we compare various ways of fusing RGB and our flow representation, shown in Fig. \ref{fig:fuse-method}. We compare no fusion, late fusion (i.e., separate RGB and flow CNNs) and addition/multiplication/concatenation fusion. In Table \ref{tab:fuse}, we compare different fusion methods for different locations in the network. We find that fusing RGB information is very important ``when computing flow directly from RGB input''. However, it is not as beneficial when computing the flow of representations as the CNN has already abstracted much appearance information away. We found that concatenation of the RGB and flow features perform poorly compared to the others. We do \textbf{not} use two-stream fusion in any other experiments, as we found that computing the representation flow after the 3rd residual block provides sufficient performance even without any fusion.

\begin{figure}
    \centering
      \includegraphics[width=\linewidth]{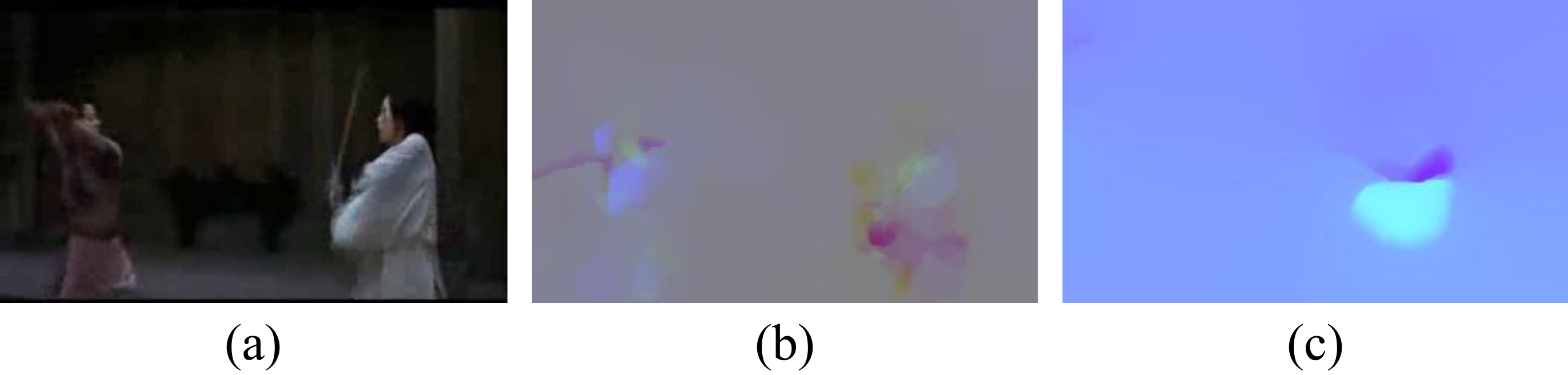}
      \caption{Example (a) RGB image, (b) TVL-1 flow image and (c) TVL-1 applied twice (i.e., Flow-of-Flow). Directly computing flow-of-flow results in poor input, as the inputs of magnitude and directions do not follow the brightness consistency assumption.}
      \label{fig:fof-ex}
\end{figure}
\begin{figure}
    \centering
      \includegraphics[width=\linewidth]{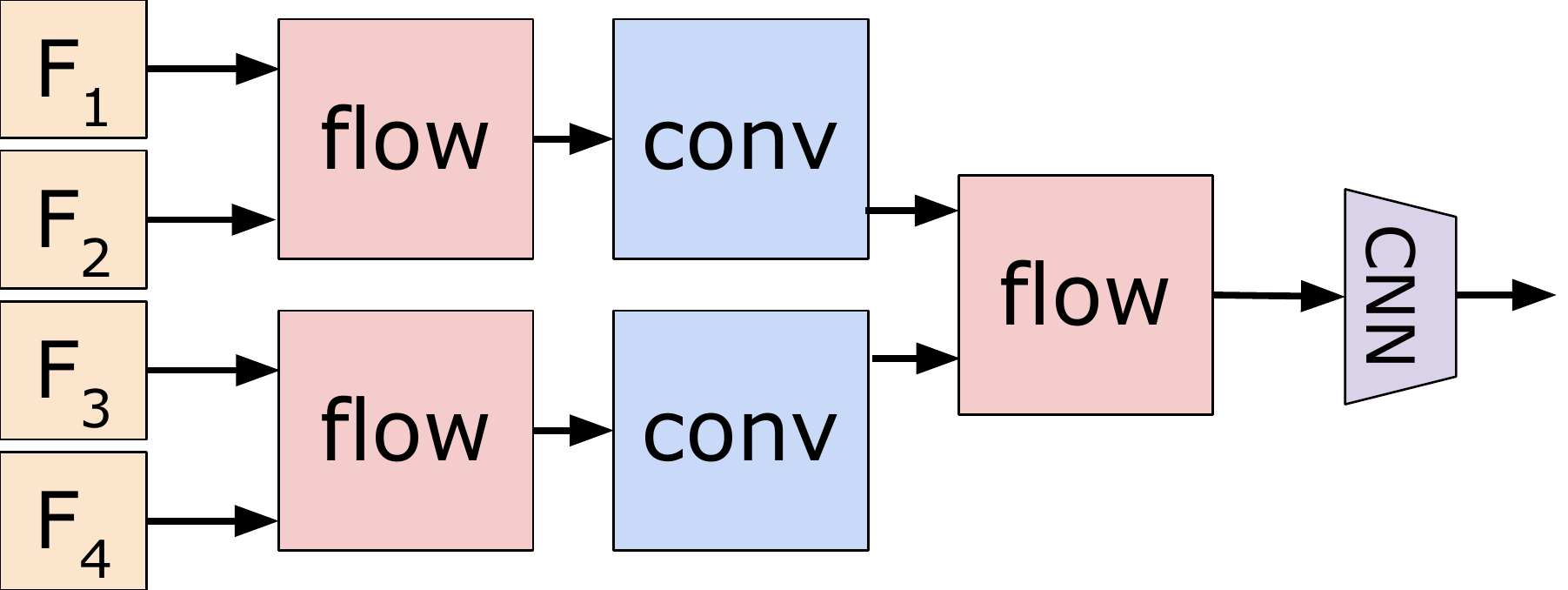}
      \captionof{figure}{Illustration of how our model computes the FoF. Adding the intermediate conv layer allows for the smoothing of flow and conversion from magnitude+direction to feature values. This allows a second flow layer to further refine the motion feature.}
      \label{fig:fof-model}
\end{figure}

\begin{table}
  \caption{Different fusion methods for flow computed at different locations in the network on our Tiny-Kinetics dataset using 10 iterations with flow parameter learning.}
  \label{tab:fuse}
  \centering
  \begin{tabular}{lccc}
    \toprule
                      &  RGB & 1 Block & 3 Blocks \\
    \midrule
    None              & 37.4  & 52.4   & 59.4 \\
    Late              & 61.3  & 60.4   & 61.5 \\
    Add               & 59.7  & 57.2   & 56.5  \\
    Multiply          & 58.3  & 58.1   & 57.8 \\
    Layer + Multiply  & 60.1  & 61.7   & 61.7 \\
    Concat            & 42.4  & 48.5   & 47.6 \\
    \bottomrule
  \end{tabular}
\end{table}

\vspace{-3pt}
\paragraph{Flow-of-flow} We can stack our layer multiple times, computing the flow-of-flow (FoF). This has the advantage of combining more temporal information into a single feature. Our results are shown in Table \ref{tab:flow-of-flow}. Applying the TV-L1 algorithm twice gives quite poor performance, as optical flow features do not really satisfy the brightness consistency assumption, as they capture magnitude and direction of motion (shown in Fig.~\ref{fig:fof-ex}). Applying our representation flow layer twice performs significantly better than TV-L1 twice, but still worse than our baseline of not doing so. However, we can add a convolutional layer between the first and second flow layer, flow-conv-flow (FcF), (Fig.~\ref{fig:fof-model}), allowing the model to better learn longer-term flow representations. We find this performs best, as this intermediate layer is able to smooth the flow and produce a better input for the representation flow layer. However, we find adding a third flow layer reduces performance as the motion representation becomes unreliable, due to the large spatial receptive field size. In Fig. \ref{fig:vis-flow}, we visualize the learned flow-of-flow, which is a smoother, acceleration-like feature with abstract motion patterns.

\begin{table}
  \caption{Computing the FoF representation. TV-L1 twice provides poor performance, using two flow layers with a conv. in between provides the best performance. Experiments used 10 iterations and learning flow parameters.}
  \label{tab:flow-of-flow}
  \centering
  \begin{tabular}{lc}
    \toprule
                                 &  Tiny-Kinetics \\
    \midrule
    TVL-1 twice                  &  12.2   \\
    Single Flow Layer            &  59.4  \\
    Flow-of-Flow                 &  47.2 \\
    Flow-Conv-Flow (FcF)         &  {\bf 62.3}  \\
    Flow-Conv-Flow-Conv-Flow     &  56.5  \\
    \bottomrule
  \end{tabular}
\end{table}

% \paragraph{Learning vs. feature extractor} Have results, could put in appendix to save space if needed.

\vspace{-3pt}
\paragraph{Flow of 3D CNN Feature} Since 3D convolutions capture some temporal information, we test computing our flow representation on features from a 3D CNN. As 3D CNNs are expensive to train, we follow the method of I3D \cite{carreira2017quo} to inflate a ResNet-18 pretrained on ImageNet to a 3D CNN for videos. We also compare to the (2+1)D method of spatial conv. followed by temporal conv from \cite{xie2017rethinking}, which produces a similar feature combining spatial and temporal information. We find our flow layer increases performance even with 3D and (2+1)D CNNs already capturing some temporal information: Tables \ref{tab:3d-cnn} and \ref{tab:2+1d-cnn}. These experiments used 10 iterations and learning the flow parameters. In these experiments, FcF was not used.

We also compared to the OFF \cite{sun2018optical} using (2+1)D and 3D CNNs. We observe that this method does not result in meaningful performance increases using CNNs that capture temporal information, while our approach does.

%that even though 3D and (2+1)D CNNs capture some temporal information, the application of the iterative representation flow layer increases performance.  

\begin{table}
  \caption{Computing representation flow using 3D ResNet-18. We find that even though 3D CNNs capture some temporal information, the use of the iterative representation flow further improves performance.}
  \label{tab:3d-cnn}
  \centering
  \begin{tabular}{lc}
    \toprule
                          &  Tiny-Kinetics  \\
    \midrule
    RGB 3D ResNet-18      & 54.6  \\
    TVL-1 3D ResNet-18    & 37.6  \\
    Two-Stream 3D ResNet  & 57.5   \\
    RGB-Only OFF \cite{sun2018optical} & 54.8 \\
    Input (RGB)           & 38.5    \\
    After Block 1         & 58.4   \\
    After Block 3         & {\bf 59.7}  \\
    \bottomrule
  \end{tabular}
  \end{table}
  \begin{table}
  \caption{Computing representation flow using (2+1)D ResNet-18. We find that the representation flow layer is beneficial with this base network, confirming it captures features standard spatio-temporal convolution does not.}
  \label{tab:2+1d-cnn}
  \centering
  \begin{tabular}{lc}
    \toprule
                          &  Tiny-Kinetics  \\
    \midrule
    RGB (2+1)D ResNet-18      & 53.4  \\
    TVL-1 (2+1)D ResNet-18    & 36.3  \\
    Two-Stream (2+1)D ResNet  & 55.6   \\
    RGB-Only OFF \cite{sun2018optical} & 53.7 \\
    Input (RGB)               & 39.2    \\
    After Block 1             & 57.3   \\
    After Block 3             & {\bf 60.7}  \\
    \bottomrule
  \end{tabular}
\end{table}

\vspace{-3pt}
\paragraph{Comparison to other motion representations} We compare to existing CNN-based motion representation methods to confirm the usefulness of our representation flow. For these experiments, when available, we used code provided by the authors and otherwise implemented the methods ourselves. To better compare to existing works, we used ($16\times$) $224\times 224$ images. Table~\ref{tab:other-motion} shows the results. MFNet \cite{lee2018motion} captures motion by spatially shifting CNN feature maps, then summing the results, TVNet \cite{fan2018end} applies a convolutional optical flow method to RGB inputs, and ActionFlowNet \cite{ng2018actionflownet} trains a CNN to jointly predict optical flow and activity classes. We also compare to OFF \cite{sun2018optical} using only RGB inputs. Note that the HMDB performance in \cite{sun2018optical} was reported using their three-stream model (i.e., RGB + RGB-diff + optical flow inputs), and here we compare to the version only using RGB. Our method, which applies the iterative flow computation on CNN feature maps, performs the best.

\begin{table}
  \caption{Comparisons to other CNN-based motion representations, using 10 iterations and learning flow parameters. This is without FcF and two-stream fusion.}
  \label{tab:other-motion}
  \centering
  \begin{tabular}{lcc}
    \toprule
                                                &  Tiny-Kinetics & HMDB  \\
    \midrule
    ActionFlownet \cite{ng2018actionflownet}   & 51.8 & 56.2 \\
    MFNet \cite{lee2018motion}                 & 52.5 & 56.8 \\
    TVNet \cite{fan2018end}                    & 39.4 & 57.5 \\
    RGB-OFF \cite{sun2018optical}              & 55.6 & 56.9 \\
    Ours                                        & {\bf 61.1} & {\bf 65.4} \\
    \bottomrule
  \end{tabular}
\end{table}
%Ours & 61.1 & 65.4

\begin{table*}
  \caption{Comparison to the state-of-the-art action classifications. `HMDB(+Kin)' means that the model was pre-trained on Kinetics before training/testing with HMDB. Missing results are due to those papers not reporting that setting. We marked the best performances (per dataset) with bold texts. Note that all our models have a single-stream design.}
  \label{tab:sota}
  \centering
  \begin{tabular}{lcccc}
    \toprule
                     &  Kinetics & HMDB & HMDB(+Kin) & Run-time (ms)  \\
    \midrule
    \textbf{2D CNNs}\\
    RGB               & 61.3     & 53.4 &    59.4    & 225 $\pm 15$ \\
    Flow              & 48.2     & 57.3 &    61.2    & 8039 $\pm 140$ \\
    Two-stream        & 64.5     & 62.4 &    66.6    & 8546  $\pm 147$\\
    TVNet (+RGB) \cite{fan2018end}     & -    &  71.0 & - & 785 $\pm 21$ \\
    OFF (RGB Only) \cite{sun2018optical}     & -    &  57.1 & - & 365 $\pm 26$ \\
    OFF (RGB + Flow + RGB Diff) \cite{sun2018optical}     & -    &  74.2 & - & 9520 $\pm 156$ \\
    Ours (2D CNN + Rep. Flow)      & 68.5 &  73.5 & 76.4 & 524 $\pm 24$ \\
    Ours (2D CNN + FcF)        & 69.4 &  74.4 & 77.3 & 576 $\pm 22$ \\
    \midrule
    \textbf{(2+1)D CNNs} \\
    % P3D \cite{qiu2017learning}    not tested on these datasets.. 
    RGB R(2+1)D \cite{tran2018closer}        & 74.3 & -  & 74.5  & 471 $\pm 18$ \\
    Two-Stream R(2+1)D \cite{tran2018closer} & 75.4 & -  & 78.7  & 8623 $\pm 152$ \\
    Ours ((2+1)D CNN + Rep. Flow)    & 75.5 & -     & 77.1 & 622 $\pm 23$ \\
    Ours ((2+1)D CNN + FcF)    & 77.1 & -     & \textbf{81.1} & 654 $\pm 21$ \\
    Ours ((2+1)D CNN + FcF) + Non-local   & \textbf{77.9} & -     & \textbf{81.1} & 865 $\pm 21$ \\
    \midrule
    \textbf{3D CNNs} \\
    RGB S3D \cite{xie2017rethinking}         & 74.7 & -   &  75.9 & 525 $\pm 22$ \\
    Two-Stream S3D \cite{xie2017rethinking}  & 77.2 & -   &  -   & 8886  $\pm 162$ \\
    I3D (RGB)  \cite{carreira2017quo}       & 71.1 & 49.8  & 74.3 & 594  $\pm 23$ \\
    I3D (Flow)        & 63.4 & 61.9 & 77.3 & 8845  $\pm 148$ \\
    I3D (Two-Stream)    & 74.2 & 66.4 & 80.7 & 9354  $\pm 154$ \\
    ResNet-101 + Non-local \cite{wang2017non} & 77.7 & - & - & 3750 $\pm 125$\\
    \bottomrule
  \end{tabular}
\end{table*}

\vspace{-3pt}
\paragraph{Computation time} We compare our representation flow to state-of-the-art two-stream approaches in terms of run-time and number of parameters.
%Run-time is reported using the OpenCV TVL-1 optical flow algorithm on a GPU. 
All timings were measured using a single Pascal Titan X GPU, for a batch of videos with size $32\times 224\times 224$. 
The flow/two-stream CNNs include the time to run the TV-L1 algorithm (OpenCV GPU version) to compute the optical flow. All CNNs were based on the ResNet-34 architecture. As also shown in Table \ref{tab:sota}, our method is significantly faster than two-stream models relying on TV-L1 or other optical flow methods, while performing similarly or better. The number of parameters our model has is half of its two-stream competitors (e.g., 21M vs. 42M, in the case of 2D CNNs).

%\begin{table}
%  \caption{Comparison of the run-time and number of parameters of various methods for a single video.}
%  \label{tab:complexity}
%  \centering
%  \begin{tabular}{lcc}
%    \toprule
%                            & Run-time (ms) & Number Parameters  \\
%    \midrule
%    RGB 2D CNN              &     225       &  21M \\
%    TVL-1 + Flow 2D CNN     &     8039      &  21M \\
%    Two-stream 2D CNN       &     8546      &  42M \\
%    Ours                    &     524       &  21M \\
%    \bottomrule
%  \end{tabular}
%\end{table}

\vspace{-3pt}
\paragraph{Comparison to state-of-the-arts} We also compared our action recognition accuracies with the state-of-the-arts on Kinetics and HMDB. For this, we train our models using $32\times 224\times 224$ inputs with the full kinetics dataset, using 8 V100s. We used the 2D ResNet-50 as the architecture. Based on our experiments, we applied our representation flow layer after the 3rd residual block, learned the hyperparameters and divergence kernels, and used 20 iterations. We also compare our flow-of-flow model. Following \cite{szegedy2016rethinking}, the evaluation is performed using a running average of the parameters over time. Our results, shown in Table \ref{tab:sota}, confirm that this approach clearly outperforms existing models using RGB only inputs, and is competitive against expensive two-stream networks. Our model performs the best among those not using optical flow inputs (i.e., among the models only taking $\sim$600ms per video). The models requiring optical flow were more than 10 times slower, including two-stream versions of \cite{carreira2017quo,wang2017non,xie2017rethinking}

\section{Conclusion}
We introduced a learnable representation flow layer inspired by optical flow algorithms. We experimentally compared various forms of our layer to confirm that the iterative optimization and learnable parameters are important. Our model clearly outperformed existing methods in both speed and accuracy on standard datasets. We also introduced the concept of `flow of flow' to compute longer-term motion representations and showed it benefits performance.

\paragraph{Acknowledgement} This work was supported in part by the National Science Foundation (IIS-1812943 and CNS-1814985).

{\small
\bibliography{bib}
\bibliographystyle{ieee}
}

\clearpage
\newpage
\appendix
\section{Training and Implementation Details}
\label{app:train}

\paragraph{Implementation Details} When applying the representation flow layer within a CNN, we first applied a 1x1 convolutional layer to reduce the number of channels from $C$ to 32. CNN feature maps often have hundreds of channels, but computing the representation flow for hundreds of channels is computationally expensive. We found 32 channels to be a good trade-off between performance and speed. The flow layer produces output with 64 channels, $x$ and $y$ flows for the 32 input channels, which are concatenated together. We apply a 3x3 convolutional layer to this representation to produce $C$ output channels. This allows us to apply the rest of the standard CNN to the representation flow feature.

Two-stream networks stack 10 optical flow frames to capture temporal information \cite{simonyan2014two}. However, we found that stacking representation flows did not perform well. Instead, we computed the flow for sequential images and averaged the predictions from a sequence of 16 frames. We found this outperformed stacking flow representations.

\paragraph{Training Details} We trained the network using stochastic gradient descent with momentum set to 0.9. For Kinetics and Tiny-Kinetics, the initial learning rate was 0.1 and decayed by a factor of 10 every 50 epochs. The model was trained for 200 epochs. The 2D CNNs were trained using a batch size of 32 on 4 Titan X GPUs. The 3D and (2+1)D CNNs were trained with a batch size of 24 using 8 V100 GPUs. When fine-tuning on HMDB, the learning rate started at 0.005 and decayed by a factor of 10 every 20 epochs. The network was fine-tuned for 50 epochs. When learning the optical flow parameters, the learning rate for the parameters (i.e., $\lambda, \tau, \theta$, divergence kernels and Sobel filters) was set of $0.01\cdot\text{lr}$, otherwise the model produced poor predictions. This is likely due to the accumulation of gradients from the many iterations of the algorithm. For Kinetics and Tiny-Kinetics, we used dropout at 0.5 and for HMDB it was set to 0.8.

\paragraph{Testing Details} For the results reported in Table \ref{tab:sota}, we classified actions by applying our model to 25 different random croppings of each video. As found in many previous works, this helps increase the performance slightly. In all the other experiments (i.e., Tables \ref{tab:where}-\ref{tab:other-motion}), random cropping was not used. Also notice that only the results in Table \ref{tab:sota} uses our full model with $32\times 224\times 224$ input resolution. The other experiments uses spatially and/or temporally smaller models.

\end{document}

%% file: math_commands.tex
\usepackage{amsmath,amsfonts,bm}

\def\eqref#1{equation~\ref{#1}}
\def\1{\bm{1}}

\DeclareMathAlphabet{\mathsfit}{\encodingdefault}{\sfdefault}{m}{sl}
\SetMathAlphabet{\mathsfit}{bold}{\encodingdefault}{\sfdefault}{bx}{n}